\documentclass[11pt]{article}
\usepackage[utf8]{inputenc}
\usepackage[english]{babel}
\usepackage{amsmath}
\usepackage{emptypage}
\usepackage{mathtools}
\usepackage{amssymb}
\usepackage{graphics}
\usepackage{enumitem}
\usepackage{amsthm}
\theoremstyle{plain} 
\newtheorem{theorem}{Theorem}[section]

\theoremstyle{remark}

\theoremstyle{definition}

\theoremstyle{plain} 

\theoremstyle{plain}

\usepackage[left=1.5in, right=1in, top=1in, bottom=1in, includefoot, headheight=13.6pt, twoside]{geometry}

\title{An alternative formulation of attention pooling function in translation}
\author{Eddie Conti}
\date{July 2024 \\
econtico8@alumnes.ub.edu}

\begin{document}

\maketitle
\section*{Abstract}
\small{
The aim of this paper is to present an alternative formulation of the attention scoring function in translation tasks. Generally speaking, language is deeply structured, and this is reflected in the attention scoring matrix (see Figure \ref{table_approx} and Figure \ref{table2}). We exploit this property to define the attention pooling function, taking this aspect into account. In the first chapters, we introduce the attention mechanism in mathematical terms and explain its limitations and alternative formulations. Next, we focus on the experimental session that led to the alternative formulation. Essentially, we guide queries and keys to interact in a specific manner, encoding the distinct roles of attention heads and directing values on where to seek context. In mathematical terms, we can think of this formula as projecting the attention scores matrix, say $H$, onto the space of band matrices with fixed bandwidth. This convex subspace is clearly finite-dimensional and therefore closed. As a consequence, the projection on this space is well-posed and unique. However, at the price of losing the uniqueness of the projection (i.e., the best approximation for $H$), we defined a new space consisting of band matrices plus error sparse matrices. We prove that this is a compact subspace which guarantees the existence of a matrix that best approximates $H$. We conclude the thesis by validating the new formula, namely calculating how well the new formula for attention scores approximates the original one. Additionally, we explore the impact of different parameters such as w (context windows) and num-pos (number of relevant words in a sentence). These analyses provide deeper insights into how languages are processed and translated, revealing nuances in the roles of context and word relevance.
\vskip2mm \noindent
All the experiments, code, images and papers can be found in the following GitHub repository: https://github.com/EddieConti/Master-Thesis-UB}
\normalsize{}
%\textbf{AMS Subject Classification:} 68T01, 68T07. \\
%\textbf{Key Words:}  Transformers, Deep Learning, Machine Learning, Computer Science, Hilbert Spaces, optimization problems, Attention mechanism.
\section*{Acknowledgments}
I would like to make a special thanks to professors Arturo Vieiro Yanes and Oriol Pujol Vila from Universitat de Barcelona (UB) for their assistance and support during the writing of the article.

\section{Introduction}
Large Language Models (LLMs) are AI systems capable of understanding and generating human language by processing vast amounts of text data. In recent years, specifically from $2017$ with paper \cite{Mainarticle}, the use of LLMs significantly increased thanks to the introduction of the Transformer architecture. The core aspect of this architecture is the Attention Mechanism, able to simulate how human attention works by assigning varying levels of importance to different words in a sentence considering the relationship within words and their meaning.  \vskip2mm 
\noindent
In mathematical terms it can be formalized as it follows:
given $m \in \mathbb{N}$ pairs of key and value vectors $(k_i,v_i)$, $k_i,v_i \in \mathbb{R}^{d}$ where $d$ is the dimension of the embedding, for a given query vector $q_j \in \mathbb{R}^{d}$ the attention mechanism computes an output vector $o_j$ as
\[
o_j=\sum_{i=1}^{m} \alpha(q_j,k_i)g(v_i).
\]
The function $g(\cdot)$ is a linear application from $\mathbb{R}^{d}$ to $\mathbb{R}^{d}$, while function $\alpha(\cdot,\cdot)\,\colon \mathbb{R}^{d}\times \mathbb{R}^{d} \to \mathbb{R}^{+}$ computes the attention weights. For instance,
in the case of transformers,
\[
\alpha(q_j,k_i)=\frac{exp(\langle q_j,k_i\rangle)}{\sum_{l=1}^{m} exp(\langle q_j,k_l\rangle)}, \qquad g(v_i)=v_i.
\]
In literature, the function $\alpha$ is generally referred to as \textit{attention pooling} and it can be thought as a similarity measure. In general we can ask $\alpha$  to have various properties, but in general, for a fixed query vector $q_j$, we require the weights $\alpha(q_j,k_i)$ to form a convex combination, i.e., 
\[
\sum_{i=1}^{m} \alpha(q_j,k_i)=1, \quad \alpha(q_j,k_i)\geq 0 \qquad j=1,\ldots,m.
\]
To understand its meaning let us consider function $\alpha$ to be the normalized dot product,
\[
o_j= \frac{1}{C} [ \langle q_j,k_1\rangle g(v_1)+\ldots+\langle q_j,k_m\rangle g(v_m)], \quad C=\sum_{i=1}^{m} \langle q_j,k_i\rangle.
\]
Now if, for example, $q_j$ attends mostly $k_1$, then
\[
o_j \approx \frac{1}{C}[\langle q_j,k_1 \rangle g(v_1)] \approx g(v_1).
\]
Therefore, we can think of attention weights as a ``guide map" to values vectors $g(v_i)$. 
\vskip2mm 
\noindent
\subsection{Limitations of Attention Mechanism and alternative formulations}
The Transformer architecture suffers from limitations when it comes to processing long sequences. Most of the complexity of the architecture lies in the self-attention mechanism, where we compute the attention coefficients. Each token $x_i$ is related to every other token $x_j$ in order to capture the relationship between words. This means that the resulting complexity is $O(n^2)$, where $n$ is the sequence length. As a consequence, several attempts have been made in literature to tackle this challenge (see \cite{bert}, \cite{bigbird}, \cite{longform}, \cite{approx}, \cite{linear transformer}, \cite{matrix approx}). \vskip2mm \noindent
For instance, in the case of BigBird, the generalized attention for head $h$ is defined as 
\begin{equation} \label{sparse_att}
\text{ATTN}_D(X)_i=x_i+\sum_{h=1}^H\sigma\Bigl(Q^{(h)}(x_i)K^{(h)}(X_{N(i)})^T\Bigr)\cdot V^{(h)}(X_{N(i)}),
\end{equation}
where $Q^{(h)},K^{(h)}: \mathbb{R}^d \to \mathbb{R}^k$ ($k$ is the dimension of each head) are query and key functions, $V_h: \mathbb{R}^d \to \mathbb{R}^d$ is a value function, $\sigma$ is a scoring function such as softmax and $H$ denotes the number of heads. Denote $X_{N(i)}$ the matrix formed by only stacking $\{x_j : j \in N(i)\}$ and not all the inputs. It is clear that \eqref{sparse_att} reduces to the usual formulation (see paper \cite{Mainarticle}) if $N(i)$ is full, in the sense that we are considering the relationship among all the input words. The general idea behind BigBird architecture is that most contexts within NLP and computational biology have data which displays a great deal of locality of reference. \vskip2mm \noindent
Alternatively, as for paper \cite{matrix approx}, attention values for a given head $i$ are computed as 
\begin{equation} \label{linformer_structure}
    head_i=\underbrace{Softmax\Bigl(\frac{Q^{(i)}(E_iK^{(i)})^T}{\sqrt{k}}\Bigr)}_{n \times k'}\underbrace{F_iV^{(i)}}_{k'\times d}.
\end{equation}
In equation \eqref{linformer_structure} we project the $(n\times k)$-dimensional key and values layers $K^{(i)}$, $V^{(i)}$ into $(k'\times k)$-dimensional projected key and value layers. Thus, if we can choose a very small projected dimension $k'$, then we can significantly reduce the memory and space consumption.

\section{Experiments on attention layer}
In this section we are going to present our results from the experiments that will lead to an alternative formulation for the attention weights. It is important to underline a couple of aspects before continuing: the aim of this paper is to describe and analyze Self-Attention in mathematical terms. As a consequence, the experiments with the Transformer architecture are not optimal and we trained, for computational time reasons, just using one transformer block. For instance the results we got from the translation task are very far from the proper translation.
Moreover, we focused on the original functions
\[
\alpha(q_j,k_i)=\frac{ \exp(\langle q_j,k_i\rangle)}{\sum_{l=1}^{m} \exp(\langle q_j,k_l\rangle)}, \qquad g(v_i)=v_i.
\]
The conclusions we draw may be different with other functions. 

\vskip2mm \noindent
In the experimental part we used the opus\_books from HuggingFace from English to Italian\footnote{see https://huggingface.co/docs/datasets/index}, which included a large variety of texts, articles and web pages. In order to understand at a basic level what is the attention layer doing, we decided to train the transformer for $20$ epochs with just $1$ transformer block and either $1$ or $8$ heads, to better understand the flow of information and prevent the depth of the neural network from altering the scores. The goal is to produce a correct translation of the sentence involved, so the attention scores reflect this task.
\begin{figure}[!h]
    \centering
    \includegraphics[scale=0.5]{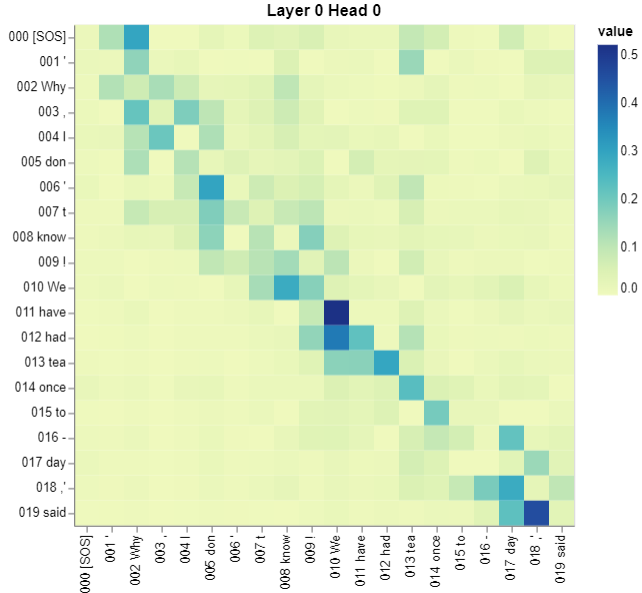}
    \caption{Attention visualization architecture with 1 layer and 1 head}
    \label{attn1.1}
\end{figure}
\begin{figure}[!h]
    \centering
    \includegraphics[scale=0.27]{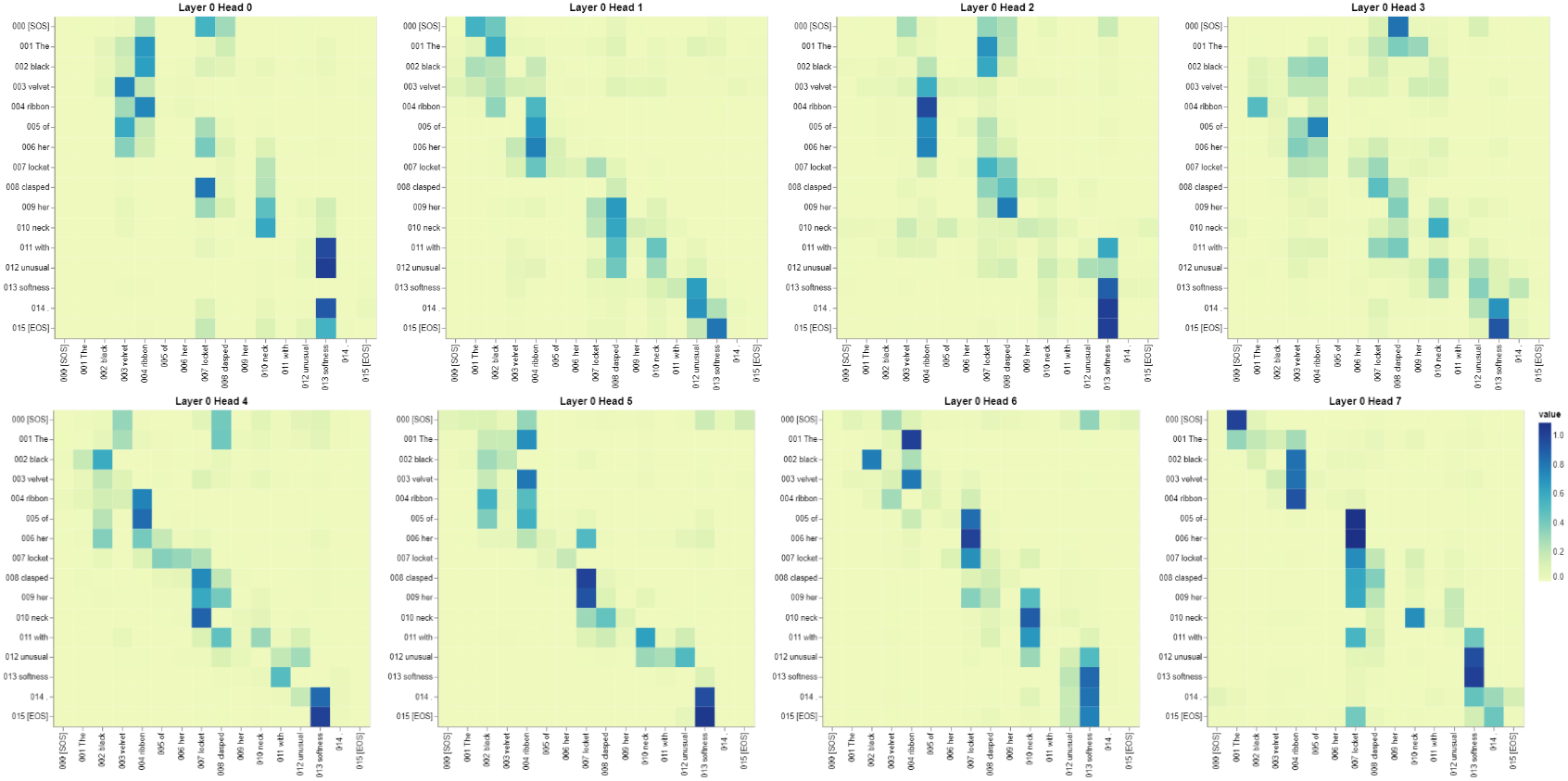}
    \caption{Attention visualization architecture with 1 layer and 8 heads}
    \label{attn1.8}
\end{figure} \\ \noindent
Before describing the above figures, let us remark the fact that, in paper \cite{elena article} the matrices are better structured (and we will explain what do we mean by structure) as their experiments involved a more complex architecture. Nevertheless, our experiments, with a certain degree of error, still confirm their results. \\
After adapting the code of the transformer architecture we used a function to get and visualize the attention matrix for a specific sentence. We then focused on visually analysing the structure of the arrays for many different examples and architectures: with $1$ head, with $8$. In Figure \ref{attn1.1} and \ref{attn1.8} we report an example of our experimental session. These figures clearly show that attention weights matrices exhibit a certain structure while performing translation. Most of the heads show a diagonal structure in the matrix since translating the token referencing to the word itself should be the most important to produce the translated token. Furthermore, it is possible to see in head $2$ and head $7$ that there is stress on a specific token and this is identified by a darker vertical column in the matrix. This phenomenon captures another aspect of translating: in producing language there are words that are giving the general meaning or context of the whole sentence or words that are referencing to other words such as pronouns. As a consequence, these layer with token attending a specific one are encoding this specific aspect. Furthermore it is worthy to underline that all the matrix involved are full rank with low $l_2$ norm. 
\subsection{On the structure of attention weights}
The experiments carried shows a specific structure in the attention pooling function for the transformer
\[
f(Q,K)=Softmax\Bigl(\frac{QK^T}{\sqrt{k}}\Bigr).
\]
Three general structured heads can be identified:
\begin{itemize}
    \item positional heads: tokens attend to a token's immediate neighbors;
    \item syntactic heads: tokens attending another token because of linguistic relationship;
    \item rare tokens heads: a group of token attending a specific token, usually an infrequent token.
\end{itemize}
What is interesting is that the attention scores are never spread among the matrix but the values are mostly concentrated around the diagonal, up to a proper shift of them. Taking into consideration this and the roles of heads we propose the following formulation for the attention scores:
\begin{equation} \label{new_scores}
    f(P,\tilde{E})=(I_nP+\tilde{E}), \qquad P,\tilde{E} \in \mathbb{R}^{n\times n}, \quad P \in \Sigma \subset\mathcal{M}_n(\mathbb{R})
\end{equation}
and therefore the new attention values
\[
 (I_nP+\tilde{E})V.
\]
We write $I_nP$ instead of $P$ to stress on the action of $P$ in shifting the attention. The matrix $\tilde{E}$ is an sparse error matrix with elements $|\epsilon_{i,j}|\leq \epsilon$ so with very low values, while $P$ is a matrix taken from a set $\Sigma$ with a specific structure that we will discuss afterwards. Before describing the matrices we are dealing with and their properties let us clarify the intuition behind formula \eqref{new_scores}: the role of the matrix $P$ is to shift the positional heads, i.e. the diagonal, to the relevant structure that attention scores exhibit. In some sense, we are forcing query and keys to attend each other in a specific form such that encode the different roles of attention heads and so telling values where to look to get the context. After doing that we add an error matrix to introduce randomness and add elasticity to the formula. 
\begin{figure}[!h]
    \centering
    \includegraphics[scale=0.6]{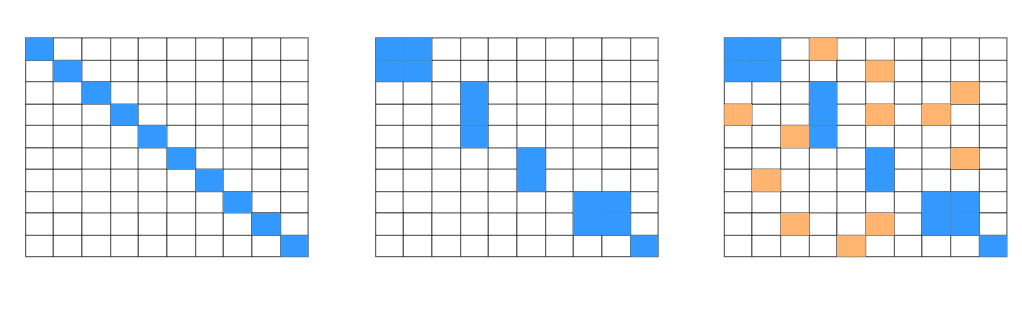}
    \caption{\small{Effect of shifting the attention and adding sparse error. We start from the identity, we apply (central image) the matrix $P$ that is shifting the attention and creating a specific structure, then (right image) we add some random noise, represented by orange blocks, to add robustness to our formula.}}
    \label{my_att}
\end{figure} \\
For example in Figure \ref{my_att} we are forcing token $3,4,5$ to attend token $4$ as we can see in the central image since we detect a column of blue blocks. The matrix $P$ in \eqref{new_scores} has a specific structure. We can define $3$ family matrices that we are interested in: $\Sigma_1=I_n$ to capture the positional head as we are dealing with mostly diagonal matrices, $\Sigma_2(w)$ for a windows size $w$ which is made by appropriate matrices to capture syntactic heads, namely block matrices that correlate tokens to some other specific tokens as in Figure $12$ for head $6$: here token $5,6,7$, namely words 'of', 'her', 'locket' are attending the word 'locket' as we can see a blue vertical column. In the syntactic structure of the sentence, the preposition 'of' and the possessive pronoun 'her' both relate to the noun 'locket' according to the grammatical rule of English sentence construction. In this structure, 'of' indicates the possession relationship between 'her' and 'locket', so both tokens $5$ and $6$ must point to token $7$, 'locket', which represents the possessed object. \\
Lastly $\Sigma_3$ represents matrices able to codify rare tokens and so that all token are attending a specific one as, almost, for head $7$ in Figure $12$ in which we can see a vertical darker column for token $7$, meaning that attention is pooled on this specific one.
With this notation
\begin{equation} \label{perm_sets} 
    \Sigma= \Sigma_1 \cup \Sigma_2 \cup \Sigma_3
\end{equation}
and the matrix $P$ is taken among these possible sets. The windows size $w$ is important to control the size of the blocks in $\Sigma_2$ and also it is consistent with the idea that syntactic relationships as verb-adjective are mainly close. We will dedicate particular attention on the role of $w$ in the next pages. 
\subsection{Description of the sets}
\begin{figure}[!h]
    \centering
    \includegraphics[scale=0.5]{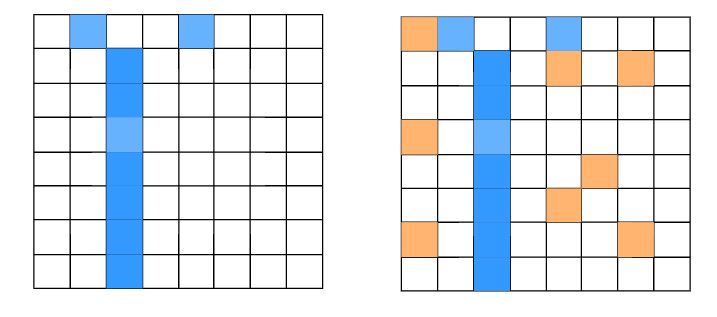}
    \caption{Exemplification of rare token heads.}
    \label{raretokenhead}
\end{figure}

\begin{figure}[!h]
    \centering
    \includegraphics[scale=0.5]{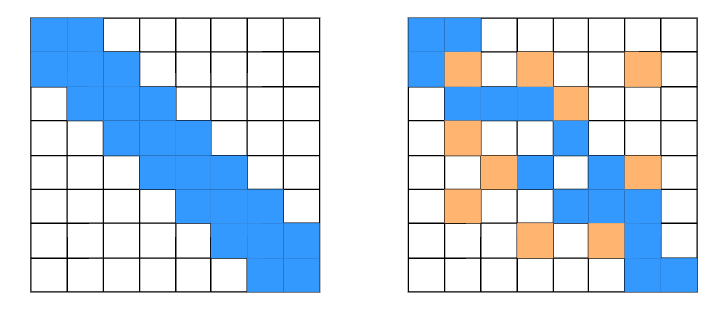}
    \caption{\small{Exemplification of syntactic heads: we start from a band matrix, turn off some blocks and then add random orange blocks.}}
    \label{syntheads}
\end{figure}

In this section we are going to describe the sets that appears in \eqref{perm_sets}. The set $\Sigma_1$ consists of $I_n$ as we are shifting the attention to the token position, and so the resulting matrix should be diagonal. The set $\Sigma_3$ is related to rare token heads of the form shown in Figure \ref{raretokenhead}
 \\
These kind of matrices are characterized by two parameters: $\mathcal{T}=\{t_1,\ldots,t_s\}$ the set of ordered tokens the attention focus on and their relative windows size $\mathcal{W}=\{w_1,\ldots,w_s\}$ in the sense that token $t_1$ is attended by $w_1$ surrounding tokens and $t_s$ by $w_s$ surrounding tokens. It is sufficient to describe a single token with its window $(t_1,w_1)$ as the general case would be just a concatenation of matrices. The base case, assuming $t_1$ in position $i$ is represented by matrices of the form
\[ 
\left[ 
\begin{array}{c|c|c} 
I_{d_1} & 0 & 0 \\ 
\hline
0 & R & 0  \\ 
\hline
0 & 0 & I_{d_2}  
\end{array} 
\right] 
\] 
where matrix $R$ is of the form:
\[
\begin{pmatrix}
    0 & \cdots & 0 & 1 & 0 & \cdots & 0 \\
    0 & \cdots & 0 & 1 & 0 & \cdots & 0 \\
    \vdots & \vdots & \vdots & \vdots & \vdots & \vdots & \vdots \\
    0 & \cdots & 0 & 1 & 0 & \cdots & 0 \\
\end{pmatrix}
\]
where we have $w$ vertical $1$s in one column $i$. This is capturing that this group of tokens are attending a specific token, pooling all attention on it. 
The base case it is just a concatenation of $I_{d_1},R,I_{d_2}$ such that $d_1+w_1+d_2=n$ in the sense that we have the identity, then the rare token block and the identity. \vskip2mm 
\noindent
Finally, the last set is $\Sigma_2$ in which we a structure close to a block diagonal matrix in which we have word, or group of words, attending other group of words capturing syntactic relationship. From the experiments it is possible to see (for example Figure \ref{attn1.8} head $4$) that group of words attending other group of words are usually close, at least in English vocabulary, therefore an easy way to describe these matrices would be to start with a multi diagonal matrix and apply a ``dropout" to obtain the desired structure (see Figure \ref{syntheads}).
 \\
The dropout can be applied also to create ``separate groups" capturing that we have no relevant information as punctuation. In mathematical terms, given a matrix $B \in \mathbb{R}^{n \times n}$ such that
\[
b_{i,j}=0 \quad \text{if}\,\, i>j+w, \quad \text{or} \quad j>i+w; \quad w\geq0
\]
where $w$ is the bandwidth, then 
\[
P=\text{dropout}(B,p), \quad \text{for some}\,\,p \in (0,1).
\] 
The three sets of matrices we defined can be thought as band matrices with an error matrix. Indeed, for rare token matrices of the form if we fix the windows size $w$, then the matrix is contained in a band matrix because the vertical column in matrix $R$ does not, by construction, exceed the bandwidth. 
\vskip2mm \noindent
We are now ready to express our ideas in mathematical and formal terms. First of all, let us define the following set for a fixed bandwidth size $w \in \mathbb{N}$:
\[
\mathcal{B}= \{ B \in \mathcal{M}_n(\mathbb{R})\,|\,\, B\,\, \text{is a $w$-band matrix}\}.
\]
Let us also consider the entrywise $1-$norm on the set of real matrices 
\begin{equation} \label{mynom}
    |M|=\sum_{i,j} |m_{i,j}|, \quad m_{i,j}\in M, \quad M \in \mathcal{M}_n(\mathbb{R}).
\end{equation} \noindent
It is clear now that $\mathcal{M}_n(\mathbb{R})$ with this norm, becomes a normed vector space. Moreover, $\mathcal{B}$ is a finite dimensional subspace of $\mathcal{M}_n(\mathbb{R})$ because if $B_1,B_2\in \mathcal{B}$ and $\lambda_1,\lambda_2 \in \mathbb{R}$ then since we have a fixed bandwidth clearly 
\[
\lambda_1B_1+\lambda_2B_2 \in \mathcal{B}.
\]
It is a known result that any finite dimensional subspace of a normed vector space is closed, hence we conclude that $\mathcal{B}$ is closed. It is also trivial that any linear subspace is also convex. In our case the space $\mathcal{B}$ has dimension
\[
n+2(n-1)+\ldots+2(n-w)=2\sum_{i=0}^{w} (n-i)-n = -(w+1)(w-2n)-n.
\]
As a consequence, we can guarantee that given the matrix $H$ for a certain head, there exists a unique $\tilde{B} \in \mathcal{B}$ such that
\begin{equation*} 
    \tilde{B} = \underset{B \in \mathcal{B}}{\text{argmin}}\,\,|H - B|
\end{equation*}
in view of the following theorem (see \cite{Damianone})
\begin{theorem}
    Let $\mathcal{H}$ be an Hilbert space with respect the norm $||\cdot ||$. If $C$ is a convex, closed, non-empty subset of $\mathcal{H}$ and $q$ a point of $\mathcal{H}$. Then, there exists a unique $p \in C$ such that
    \[
    ||p-q||=dist(C,q):=\underset{x \in C}{\text{inf}}\,\, ||x-q||.
    \]
\end{theorem} \noindent
In other words, given an attention weights matrix $H$, we can project onto the subspace $\mathcal{B}$ and obtain the best approximation of $H$ in terms of band matrices. The experiments we carried out, show that most of the information is around the diagonal, which further motivates the choice of band matrices. However, the other entries with less weight (i.e. with a lower value, the yellow blocks in Figure \ref{attn1.1} and \ref{attn1.8}) are non zero. The idea is to capture them with sparse random matrix as shown when we added the orange blocks in Figure \ref{raretokenhead} and Figure \ref{syntheads}. \\
Therefore, fixed $\epsilon$ a small real value, we define the following set
\[
\mathcal{X}=\{ B+E_{rr}\,\, |\,\, B \in \mathcal{B}_{[0,1]},\, E_{rr} \in \mathcal{M}_n(\mathbb{R})\}
\]
where $\mathcal{B}_{[0,1]}$ is the set of matrices $\mathcal{B}$ with entries in $[0,1]$ and $E_{rr}$ is a sparse matrix such that
\begin{equation} \label{maxcond}
    \max_{i,j} |e_{i,j}|\leq \epsilon, \quad  E_{rr}=(e_{i,j}).
\end{equation}
Now, it is straightforward that $\tilde{B} \in \mathcal{X}$ because matrix $H$ has entries in $[0,1]$ by definition of attention weights and we can take $E_{rr}$ as the zero matrix. The set $\mathcal{X}$ is clearly bounded w.r.t norm \eqref{mynom} because given $X_1,X_2 \in \mathcal{X}$, exists $L>0$, such that
\[
|X_1-X_2|<L.
\]
Indeed let $n_1$ the number of elements in the band and $n_2$ the remaining elements so that $n_1+n_2=n^2$, then 
\[
|X_1-X_2|<n_1+\epsilon n_2.
\]
Now, we are left to prove that the space $\mathcal{X}$ is closed. If we prove this, we can conclude that $\mathcal{X}$ is a compact set and the function $f \colon \mathcal{X} \to \mathbb{R}^{+}$ such that
\[
f(X)=|H-X|
\]
attains a minimum value on $\mathcal{X}$. Since $\tilde{B} \in \mathcal{X}$ this minimum value is less or equal than $|H - \tilde{B}|$ and so this is the right space to approximate our attention weights matrix. \\
To conclude that $\mathcal{X}$ is closed we have to prove that given $X_n \in \mathcal{X}$ such that
\[
X_n \to X \implies X \in \mathcal{X} \quad \text{for}\,\, n\to \infty.
\]
From the definition of the norm we are using, we have that
\[
   \lim_{n \to \infty} \sum_{i,j} |x^{(n)}_{i,j} -x_{i,j}|=0,\quad  x^{(n)}_{i,j}\in X_n, \quad x_{i,j}\in X
\]
and so, in particular,
\[ 
\lim_{n \to \infty} \max_{i,j} |x^{(n)}_{i,j} -x_{i,j}|=0 
\]
which implies the pointwise convergence
\[
\lim_{n \to \infty}  x^{(n)}_{i,j}=x_{i,j} \quad \forall i,j.
\]
Now, if we split $X_n=B_n+E^{(n)}_{rr}$ and focusing on the band part, it is immediate to conclude from the point-wise convergence that
\[
\lim_{n \to \infty} B_n
\]
is a band matrix with bandwidth less or equal that $w$. Now, let us focus on the limit on the error sparse matrix
\[
\lim_{n \to \infty} E^{(n)}_{rr}.
\]
According to our definition of norm and from the existence of $\lim_{n\to\infty} X_n$ we conclude that 
\[
\lim_{n \to \infty} e_{i,j}^{(n)}, \qquad E^{(n)}_{rr}=(e^{(n)}_{i,j})
\]
exists. Now, let us fix $i,j$ and assume the limit to be $l$. From \eqref{maxcond} we conclude that $l<\epsilon$. Furthermore, we have only two scenarios from the existence of the limit: either $e^{(n)}_{i,j}=0$ or $e^{(n)}_{i,j}\neq 0$ from a certain index $N$ such that $n>N$. However, the second case may happen for a relatively small amount of times as the matrices involved are sparse (otherwise we would violate this condition). Therefore, the first case must be the predominant one and we can conclude that 
\[
\lim_{n \to \infty} E^{(n)}_{rr}
\]
is ``mostly" $0$ and so we have an error sparse matrix.

\subsection{Validation of attention scores approximation}
\begin{figure}[!h]
    \centering
    \includegraphics[scale=0.45]{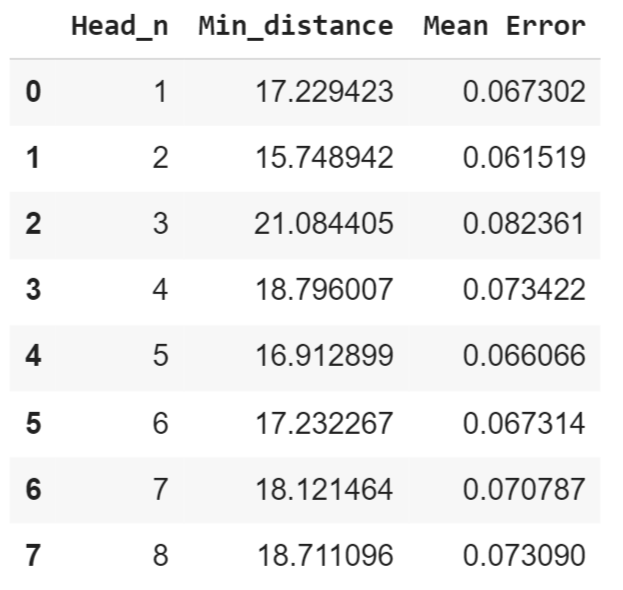}
    \caption{Validation results showing the approximation distance and the mean distance per element ($w=3$ and num-pos$=2$)}
    \label{table_approx}
\end{figure}

\begin{figure}[!h]
    \centering
    \includegraphics[scale=0.45]{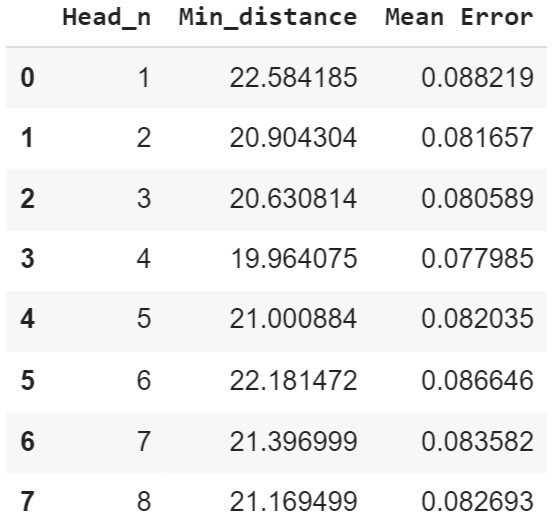}
    \caption{{Validation results showing the approximation distance and the mean distance per element ($w=10$ and $num-pos=1$)}}
    \label{table2}
\end{figure}

We conclude our project validating formula \eqref{new_scores} with respect to the original attention scores matrix. In other words, let us denote
\[
A=Softmax\Bigl(\frac{QK^T}{\sqrt{k}}\Bigr)
\]
arising after training the original transformer model, we aim at computing $|A-X|$ where $X$ is a matrix as in \eqref{new_scores}. To do so, we coded three different functions to generate the matrices belonging to sets $\Sigma_1,\Sigma_2,\Sigma_3$ and then we computed the distance between $A$ taken from the previous experiments with $8$ heads and a set of matrices generated from these sets. We took the attention score heads for a sentence of length $16$ (we can take one sentence because structure of attention scores is mostly the same, independently from the sentence taken as it can be seen in paper \cite{elena article}), and compute the minimal distance generating $30$ matrices from each $\Sigma_i$ with $w=3$ and \textit{num-pos}$=2$ (respectively the windows size/context and the number of words rare token attends). Also in this case the validation session was carried out for several matrices taken from the encoder. For simplicity we present an example of results, even tough generally they are the same, summarized in Tables \ref{table_approx} and \ref{table2}.
 \\
The Table \ref{table_approx} shows that the mean error, which is the error per entries, is relatively small, concluding that we are properly capturing the structure of the weight matrix. It is important the role of $w$ and $num-pos$. If for example we change them to be $10$ and $1$, respectively, 
we see in Table \ref{table2}, considering that the attention weights matrix is relatively small and the randomness of our approach, that the mean error significantly increases. \vskip2mm
\noindent
Each language (Italian, English,...) and/or each type of text (e.g. by author) is expected to have optimal $w$ and $num-pos$ parameters. Knowing these parameters allows to represent the attention score matrix as a matrix with structure properties that would lead to relevant reduction of the total computational cost of the LLM model.
\newpage
\section{Conclusion and further investigations}
The aim of this paper was to understand at a profound level the role of attention mechanism in transformer. We were able to provide both an analytical explanation for the formula of attention values and in particular of attention scores. The alternative and lighter formulation we found is interesting as it encodes the structure of the matrices that experiments revealed. In this scenario, then, attention is not a random process governed by back-propagation but is enriched with meaning and deeply reflects the role of translation. Far from the scope of this paper but yet interesting would be to study the role of parameters $w$ and \text{num-pos} in the code for the validation as we believe they contain valuable information on the idiom used. Indeed, by optimizing these hyperparameters for a given language we can understand how many words are relevant in a sentence and the general context size. Indeed, as pointed out in paper \cite{Context} the complexity of the language and its syntactic characteristics have to be taken into
consideration during the construction of LM architectures.

\end{document}